\documentclass{article}
\newcommand{\undertitle}[1]{}
\usepackage{arxiv,times}
\usepackage{natbib}
\setcitestyle{authoryear,round,citesep={;},aysep={,},yysep={;}}
\renewcommand{\shorttitle}{}
\date{}

\usepackage{amsmath,amsfonts,bm}

\def\eqref#1{equation~\ref{#1}}

\def\1{\bm{1}}

\DeclareMathAlphabet{\mathsfit}{\encodingdefault}{\sfdefault}{m}{sl}
\SetMathAlphabet{\mathsfit}{bold}{\encodingdefault}{\sfdefault}{bx}{n}

\usepackage{hyperref}
\usepackage{url}
\usepackage{booktabs}
\usepackage{pifont}
\newcommand{\cmark}{\ding{51}}

\usepackage{graphicx}
\usepackage{makecell}
\usepackage{multirow}
\usepackage{float}
\usepackage[table]{xcolor}

\definecolor{appendixblue}{RGB}{0,102,204}

\title{Marrying Capacity and Precision for Efficient 2D Gaussian Image Representation}

\author{
  Lin Dong\textsuperscript{1,*}, Qi Ming\textsuperscript{2,*}, Liuqian Wang\textsuperscript{1,$\dagger$}, Wentao Chao\textsuperscript{1,$\dagger$}, Yufei Gao\textsuperscript{1}, Chengming Liu\textsuperscript{1}, Lei Shi\textsuperscript{1} \\
  \textsuperscript{1}Zhengzhou University \quad \textsuperscript{2}Beijing University of Technology \\
  \texttt{\{wangliuqian, chaowentao\}@zzu.edu.cn} \\
  *Equal contribution, \quad $\dagger$Corresponding authors.
}

\begin{document}

\maketitle

\begin{abstract}
2D Gaussian representations provide an explicit and efficient way to represent images, but their rate-distortion performance under limited budgets depends critically on how Gaussian capacity and parameter precision are allocated. Existing methods typically optimize these two resources separately, overlooking content-dependent capacity demands and heterogeneous quantization sensitivity. We present CP-GS, a capacity-precision coupled Gaussian splatting framework for budget-aware 2D Gaussian image representation that jointly optimizes capacity allocation and attribute precision under a shared rate constraint. CP-GS first constructs a content-adaptive Gaussian representation by dynamically reallocating a fixed Gaussian budget from redundant regions to areas with higher representation demand. It then introduces hierarchical distortion-aware representation quantization, using efficient rendering-based calibration to allocate precision across Gaussian attributes and further refine it within heterogeneous attribute groups. Finally, it jointly optimizes Gaussian capacity and attribute precision by constructing candidate representations with different Gaussian counts and precision configurations and selecting operating points according to decoded distortion and actual serialized size. Experiments on DIV2K and Kodak show that CP-GS achieves improved rate-distortion performance over representative 2D Gaussian and implicit neural representations across a range of budgets, while maintaining a fixed overall representation budget and low optimization cost. These results demonstrate the benefit of jointly optimizing spatial capacity and parameter precision for compact Gaussian image representation.
\end{abstract}

\section{Introduction}

Efficient image representation aims to encode visual content compactly while preserving the information necessary for faithful reconstruction and downstream processing~\citep{zhang20252dmamba, kao2025bridging}. Such representations serve as a fundamental building block for a wide range of visual tasks, including image reconstruction~\citep{tian2025degradation}, super-resolution~\citep{chen2021learning}, restoration~\citep{jiang2025survey}, editing~\citep{ye2026imgedit}, and generation~\citep{xiao2025omnigen}. Building on the success of Gaussian primitives as an explicit representation for 3D scenes, 2D Gaussian splatting has recently emerged as a promising paradigm for image representation. It models visual content with a finite set of learnable anisotropic Gaussian primitives rather than dense pixels or implicit neural fields~\citep{hu2025gaussiansr}. Its explicit and differentiable formulation enables compact parameterization, efficient rendering, and flexible primitive-level manipulation, making it particularly attractive for efficient image representation. However, representation efficiency critically depends on how the model allocates Gaussian capacity and parameter precision under a constraint budget.

\begin{figure}[tbp]
\vspace{-1em}
\centering
\includegraphics[width=\textwidth]{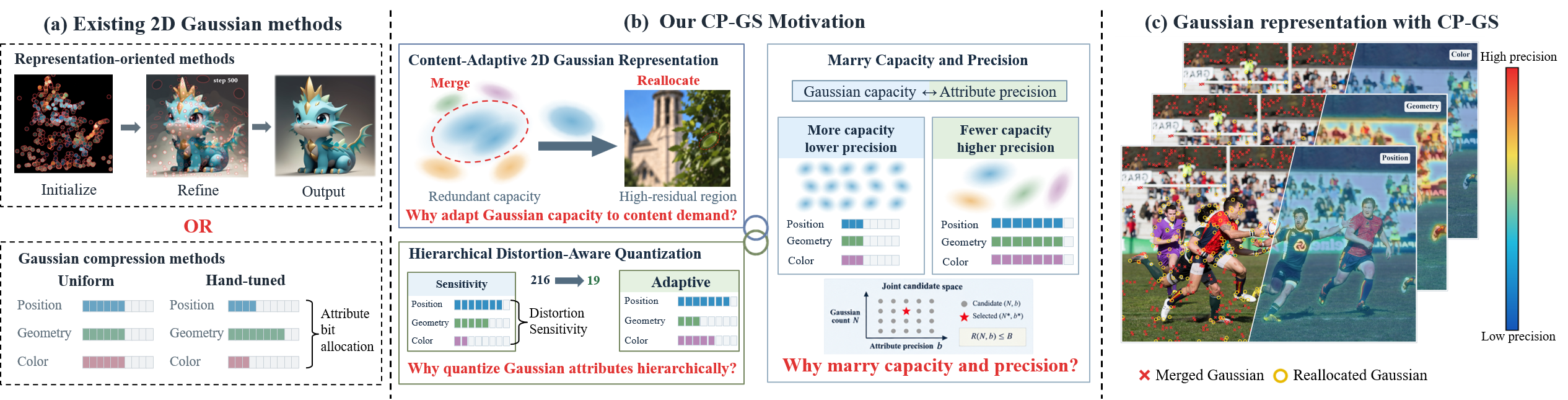}

\caption{\textbf{CP-GS at a glance.}
(a) Existing methods optimize Gaussian capacity and attribute precision separately.
(b) We identify capacity mismatch, heterogeneous precision sensitivity, and capacity-precision coupling under limited budgets.
(c) CP-GS performs content-adaptive capacity reallocation, hierarchical distortion-aware quantization, and joint capacity-precision optimization.}
\label{fig:intro}
\end{figure}

Existing approaches improve the efficiency of 2D Gaussian image representation primarily along two separate dimensions: representation capacity and parameter precision. As illustrated in Fig.~\ref{fig:intro}(a), representation-oriented methods~\citep{zhang2024gaussianimage,li2026gaussianimagepp}, such as Image-GS~\citep{zhang2025image}, focus on constructing and optimizing Gaussian primitives to match image content and reconstruction demands better. In parallel, Gaussian compression methods~\citep{liang2025structure,chen2026pcgs} reduce storage costs through uniform or manually designed quantization schemes, assigning predefined precisions to position, geometry, and color attributes. However, representation capacity and attribute precision are typically optimized separately, with the Gaussian capacity determined before attribute quantization. This decoupled treatment overlooks the fact that both resources compete for the same coding budget. Under a fixed rate, a representation may favor either more low-precision Gaussians for broader spatial coverage or fewer high-precision Gaussians for more faithful per-primitive encoding. Therefore, the preferred capacity and precision are inherently interdependent and should be jointly optimized rather than determined in isolation.

Figure~\ref{fig:intro}(b) highlights three key observations that motivate CP-GS. First, \textbf{Gaussian capacity may not match content demand.} Redundant Gaussians can accumulate in over-represented regions, while high-residual regions remain under-represented. This suggests that a fixed Gaussian budget should remain reallocatable based on content demand. Second, \textbf{Gaussian attributes exhibit heterogeneous quantization sensitivity.} Their sensitivity varies across attributes and within attributes, making uniform precision inefficient and motivating hierarchical distortion-aware quantization without exhaustive search. Third, \textbf{capacity and precision are coupled under a shared budget.} Increasing Gaussian capacity improves spatial coverage but reduces precision available to each primitive, whereas higher precision improves attribute fidelity at the cost of fewer Gaussians. This inherent trade-off motivates jointly optimizing capacity and precision in a unified rate-distortion space.

Motivated by these observations, we propose \textbf{CP-GS}, a \textbf{C}apacity-\textbf{P}recision coupled \textbf{G}aussian \textbf{S}platting framework for budget-aware 2D Gaussian image representation. CP-GS first constructs a content-adaptive representation by reallocating a fixed Gaussian budget from redundant primitives to high-demand regions. It then performs hierarchical distortion-aware quantization to allocate precision according to heterogeneous quantization sensitivity at both the attribute and intra-attribute levels. Finally, CP-GS jointly optimizes Gaussian capacity and attribute precision as competing representation resources under a shared budget, adaptively balancing spatial coverage and per-primitive fidelity. As illustrated in Fig.~\ref{fig:intro}(c), CP-GS therefore adapts both the spatial distribution and precision of Gaussian primitives to image content and representation budget. Extensive experiments on DIV2K and Kodak demonstrate that CP-GS achieves improved rate-distortion performance across a wide range of representation budgets. Our main contributions are summarized as follows:

\begin{itemize}
\item We develop a content-adaptive 2D Gaussian representation that reallocates a fixed Gaussian budget from redundant primitives to high-demand regions, improving capacity utilization without increasing the initial primitive budget.

\item We introduce hierarchical distortion-aware representation quantization that models heterogeneous quantization sensitivity at both the attribute and intra-attribute levels, enabling efficient mixed-precision allocation without exhaustive search.

\item We propose capacity-precision coupled optimization that jointly optimizes Gaussian capacity and attribute precision as competing representation resources, enabling adaptive trade-offs between spatial coverage and per-primitive fidelity under a shared budget.

\end{itemize}

\section{Related Work}

\subsection{\textbf{2D Gaussian Image Representation}}

Explicit 2D Gaussian representations have recently emerged as an efficient alternative to INR-based image codecs, such as SIREN~\citet{sitzmann2020implicit} and COIN~\citet{dupont2021coin}. Gaussian-based methods represent images with position, covariance, and color parameters, and support fast differentiable rasterization~\citep{zhang2024gaussianimage}. Subsequent works have extended this paradigm along two main directions: continuous and arbitrary-scale signal representation~\citep{hu2025gaussiansr,peng2025continuoussr}, and compact or structured encoding~\citep{pan2026sgi}. Collectively, these works demonstrate that 2D Gaussians provide an explicit and highly adaptable representation space. Our work builds on this perspective but focuses on how limited representation resources should be distributed between Gaussian capacity and parameter precision.

\subsection{\textbf{Adaptive Gaussian Modeling}}

A central problem in 2D Gaussian image representation is how to allocate a finite number of primitives according to spatially varying image complexity. Image-GS~\citet{zhang2025image} performs content-adaptive allocation by progressively placing anisotropic Gaussians according to image content. GaussianImage++~\citet{li2026gaussianimagepp} uses distortion-driven densification to allocate primitives according to signal intensity. Instant GaussianImage~\citet{zeng2025instant} predicts Gaussian parameters in a feed-forward manner, enabling self-adaptive representation without per-image optimization. LocoADC~\citet{chen2026locoadc} exploits locality in both image and Gaussian spaces through region-wise densification and similarity-driven merging. From a frequency-domain perspective, Spectrum-Adaptive Distribution~\citet{wan2025spectrum} distributes Gaussians according to spectral characteristics, while GPSToken~\citet{zhang2026gpstoken} uses entropy-driven partitioning to achieve non-uniform tokenization. At the budget level, SPARE-GS~\citet{chen2026sparegs} formulates structural evolution as a global budget-constrained optimization problem. In the 3DGS domain, GDAGS~\citet{zhou2026gdags} introduces gradient-direction-aware density control. Collectively, these methods establish adaptive density control as an effective mechanism for improving spatial capacity utilization. However, they primarily determine \emph{where} primitives should be created, retained, merged, or removed, while attribute precision is typically fixed or optimized separately. 

\subsection{\textbf{Gaussian Compression and Quantization}}

Gaussian compression reduces representation cost through quantization, pruning, and entropy coding of primitive attributes. Structure-Guided Allocation~\citet{liang2025structure} couples image structure with adaptive bitwidth quantization, while CGVQ~\citet{cheng2026cgvq} applies K-means-guided codebook quantization. In the 3DGS domain, RDO-Gaussian~\citet{wang2024rdogaussian} formulates end-to-end rate-distortion optimized compression with dynamic pruning and entropy-constrained quantization. CAT-3DGS~\citet{zhan2025cat3dgs} employs context-adaptive triplane autoregressive coding to capture inter- and intra-primitive correlations. CSGaussian~\citet{tseng2026csgaussian} combines compression and segmentation using an INR-based hyperprior, while HAC~\citet{chen2024hac} introduces hash-grid-assisted context for entropy coding. PCGS~\citet{chen2026pcgs} studies progressive compression for scalable decoding, and HilComp~\citet{huang2025hilcomp} uses Hilbert curve-based balanced clustering for compact 3D Gaussian representation. Collectively, these studies show that Gaussian attributes have heterogeneous coding requirements and that representation structure affects rate-distortion behavior. However, existing methods typically optimize precision under a fixed Gaussian capacity or treat capacity and precision as sequentially optimized resources. In contrast, CP-GS treats adaptive Gaussian modeling as one dimension of representation allocation and jointly optimizes attribute precision and Gaussian capacity under a shared budget.

\section{Preliminary}

Following prior 2D Gaussian image representations~\citep{zhang2024gaussianimage,zhang2025image}, we represent an image using a set of anisotropic Gaussian primitives defined directly on the image plane. 
Given an image $I\in[0,1]^{H\times W\times3}$, its representation with $N$ primitives is:
\begin{equation}
\mathcal G_N=
\{(\boldsymbol\mu_i,\boldsymbol\gamma_i,\mathbf c_i)\}_{i=1}^{N},
\end{equation}
where $\boldsymbol\mu_i$, $\boldsymbol\gamma_i$, and $\mathbf c_i$ denote position, geometry, and color, respectively. 
The geometry parameters $\boldsymbol\gamma_i$ define a positive-definite covariance $\boldsymbol\Sigma_i$, such that primitive $i$ provides an anisotropic spatial basis $G_i(\mathbf x)=\exp[-\frac{1}{2}(\mathbf x-\boldsymbol\mu_i)^\top \boldsymbol\Sigma_i^{-1}(\mathbf x-\boldsymbol\mu_i)]$.
A differentiable splatting renderer $\mathcal R$ aggregates the colored Gaussian responses to reconstruct the image.

Let $\mathbf b$ denote the attribute-precision configuration, yielding the quantized reconstruction $\hat I_{N,\mathbf b}=\mathcal R(Q_{\mathbf b}(\mathcal G_N))$. We characterize each representation by its distortion $D(N,\mathbf b)$ and formulate budget-aware Gaussian representation as
\begin{equation}
\min_{N,\mathbf b} D(N,\mathbf b)
\qquad
\text{s.t.}\qquad
R(N,\mathbf b)\leq B ,
\label{eq:rd_objective}
\end{equation}
where $N$ controls representation capacity, $\mathbf b$ controls attribute precision, and $B$ denotes the target budget. 

\section{Methodology}

\subsection{Overview}\label{sec:overview}

\begin{figure}[tbp]
\centering
\includegraphics[width=\textwidth]{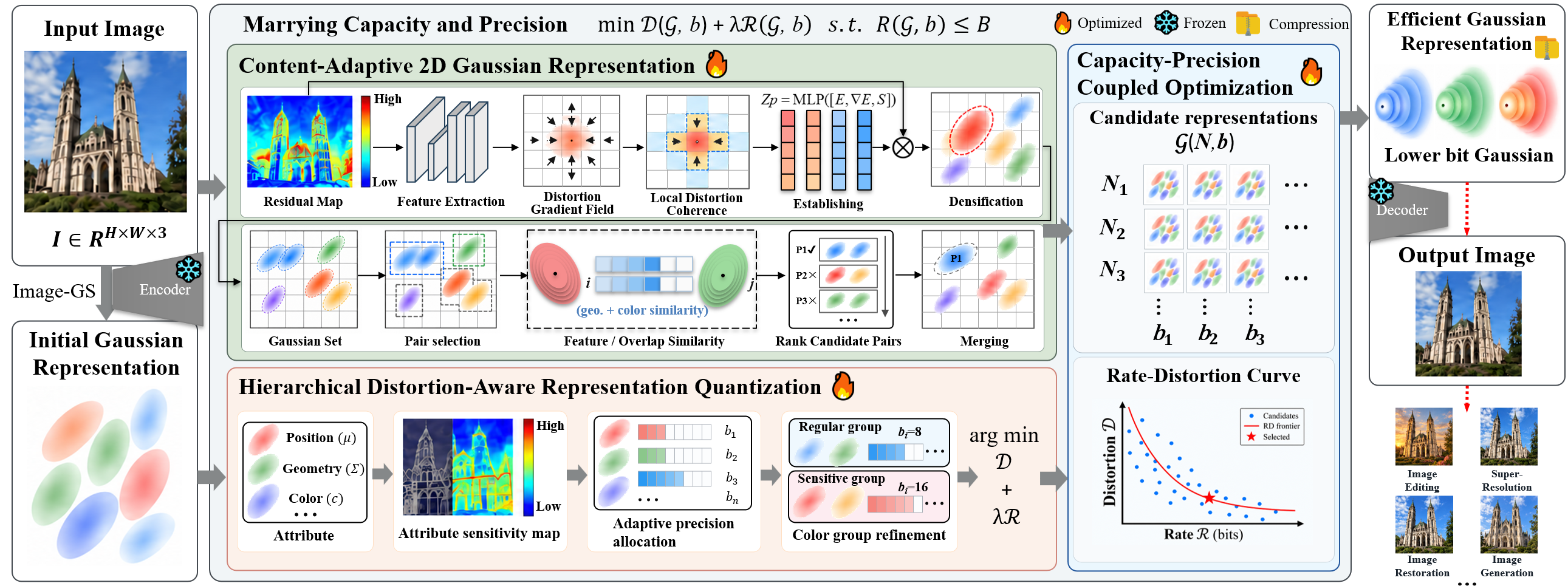}
\caption{Illustration of our CP-GS framework. Given an input image, an initial Gaussian representation is first obtained via Image-GS. Two parallel branches, content-adaptive 2D Gaussian representation and hierarchical distortion-aware representation quantization, are then jointly optimized under a shared budget, yielding a capacity–precision coupled Gaussian representation that supports efficient decoding and downstream tasks.}
\label{method}
\end{figure}

As illustrated in Fig.~\ref{method}, given an input image, CP-GS first obtains an initial Gaussian representation via Image-GS~\citep{zhang2025image}. It then applies two parallel branches: a content-adaptive 2D Gaussian representation that reallocates redundant capacity toward regions with higher reconstruction demand and a hierarchical distortion-aware representation quantization that assigns attribute precision according to rendering sensitivity and further refines heterogeneous color groups. Under a shared budget, capacity and precision configurations are jointly evaluated in a shared candidate space to select the best rate-distortion operating point, yielding an efficient and compact low-bit Gaussian representation that can be decoded for image reconstruction and downstream applications such as image editing, super-resolution, image restoration, and image generation.

\subsection{Content-Adaptive 2D Gaussian Representation}
\label{sec:spatial}

Building on the content-aware initialization of Image-GS, CP-GS improves Gaussian representation efficiency under limited budgets by merging locally redundant Gaussians and reallocating the released capacity to regions with higher representation demand.

\subsubsection{Demand-Guided Content Selection}

We estimate content demand from the reconstruction before reallocation. Let
$e(\mathbf x)=\|I(\mathbf x)-\mathcal R(\mathcal G_N)(\mathbf x)\|_2$
denote the residual magnitude. To suppress isolated errors, we measure the local coherence of residual-gradient directions:
\begin{equation}
\mathbf u(\mathbf x)=
\frac{\nabla e(\mathbf x)}
{\|\nabla e(\mathbf x)\|_2+\varepsilon},
\qquad
s_{\rm dem}(\mathbf x)=
e(\mathbf x)
\left\|
\frac{1}{|\mathcal N_{\mathbf x}|}
\sum_{\mathbf y\in\mathcal N_{\mathbf x}}
\mathbf u(\mathbf y)
\right\|_2.
\label{eq:demand}
\end{equation}

Here, $\nabla$ denotes the Sobel gradient and $\mathcal N_{\mathbf x}$ is a $7\times 7$ neighborhood. We smooth the demand map and select high-scoring local maxima using radius-three non-maximum suppression, yielding spatially separated recipient locations. Computing the demand map before merging ensures that the selected recipients reflect the original reconstruction error rather than artifacts introduced by donor removal.

\subsubsection{Redundant Gaussian Merging}

To release capacity, we identify locally stable primitives with similar appearance and overlapping spatial support. Candidate pairs are further evaluated in the rendered domain, and low-cost disjoint pairs are merged to limit reconstruction degradation. The released capacity is then reassigned to the selected high-demand locations while preserving the total number of primitives:
\begin{equation}
|\mathcal G_N^{+}|=|\mathcal G_N|.
\end{equation}
The resulting representation is subsequently refined through recovery optimization. Thus, CP-GS redistributes existing representation capacity rather than increasing the Gaussian budget. 

\subsection{Hierarchical Distortion-Aware Representation Quantization}
\label{sec:precision}

Quantization affects Gaussian attributes through different rendering mechanisms: position and geometry determine spatial support, whereas color controls appearance. CP-GS therefore performs rendered-domain distortion calibration and hierarchically allocates precision across attributes and color groups.

\subsubsection{Structured Attribute Calibration}

Let $\mathbf b=(b_P,b_G,b_C)$ denote the precision configuration for position, geometry, and color, respectively, where $b_a\in\mathcal B=\{4,6,8,10,12,16\}$. Exhaustive calibration requires evaluating $|\mathcal B|^3=216$ configurations, introducing substantial rendering overhead. Instead, CP-GS estimates attribute-wise distortion responses using a compact calibration procedure.

Let $D_{\rm ref}$ denote the distortion when all attributes are represented at 16-bit precision, and let $D_a(b)$ denote the distortion when only one attribute $a$ is quantized to $b$ bits. We approximate the joint distortion response as
\begin{equation}
\widetilde D(\mathbf b)=
D_{\rm ref}
+\sum_{a\in\{P,G,C\}}
\left[D_a(b_a)-D_{\rm ref}\right].
\label{eq:proxy}
\end{equation}

Structured-19 uses 19 logical calibration requests, corresponding to 16 unique precision configurations after caching, to efficiently approximate the finite precision search space. The selected configuration is then validated using the actual serialized representation.

\subsubsection{Hierarchical Color Precision Refinement}

Color primitives exhibit additional quantization heterogeneity within the color attribute. We therefore partition the color representation into two groups using robust channel-wise color statistics, separating primitives with atypical color responses from the dominant population. Each group is assigned its own quantization range and precision, while the selected position and geometry precisions remain fixed.

For each attribute-level operating point, we evaluate group-specific color precision candidates under the same serialized-rate constraint and select the pair that minimizes the predicted rendering distortion. The physical costs of the group map and group-specific quantization parameters are included in the final serialized representation, ensuring that the additional group structure is not treated as free side information. This hierarchical refinement captures intra-color quantization heterogeneity beyond global color precision allocation.

\subsection{Capacity-Precision Coupled Optimization}
\label{sec:joint}

Gaussian capacity and attribute precision are inherently coupled: the preferred precision depends on the retained Gaussian capacity, while the optimal capacity may change after precision adaptation. CP-GS therefore jointly optimizes these two resources within a unified rate-distortion candidate space.

\subsubsection{Budget-Conditioned Candidate Construction}

Given a fitted representation with $N_0$ primitives, we construct candidate capacities using retained ratios $r\in\{1,0.8,0.6,0.4\}$. Primitives are selected according to reconstruction residuals, local gradients, and redundancy. Precision is then recalibrated for each retained count using initial 6-, 8-, 10-, and 12-bit configurations, followed by hierarchical color refinement.

Let $\mathcal N$ denote the candidate Gaussian counts and $\mathcal A_N$ the corresponding precision configurations. The joint candidate space is
\begin{equation}
\mathcal X=
\bigcup_{N\in\mathcal N}
\{(\mathcal G_N,\mathbf b):\mathbf b\in\mathcal A_N\}.
\label{eq:candidate_set}
\end{equation}
Each candidate $x\in\mathcal X$ is evaluated using its decoded distortion $D_x$ and actual serialized size $S_x$, after which dominated candidates are removed. 

\subsubsection{Joint Capacity—Precision Optimization}

The resulting candidate pool contains representations with different combinations of Gaussian capacity and parameter precision. Here, joint optimization refers to constructing and evaluating a shared candidate space over Gaussian capacity and attribute precision, followed by rate-distortion-based selection using decoded distortion and actual serialized size. Since each candidate is physically serialized and decoded before final evaluation, selection is based on measured distortion and actual storage cost rather than estimated parameter counts.

Given a target byte budget $B$, CP-GS selects the lowest-distortion feasible representation:
\begin{equation}
x_B^\star=
\arg\min_{x\in\mathcal X:\,S_x\leq B}D_x .
\label{eq:budget_select}
\end{equation}
For continuous rate control, we further use the Lagrangian formulation.
\begin{equation}
x_\lambda^\star=
\arg\min_{x\in\mathcal X}
\left[D_x+\lambda R_x\right],
\qquad
R_x=\frac{8S_x}{HW},
\quad \lambda>0 .
\label{eq:joint_rdo}
\end{equation}

Both objectives search the same capacity-precision space but address different rate-control requirements. The hard-budget formulation selects the best representation satisfying a storage constraint while retaining non-dominated operating points missed by a linear rate penalty. More importantly, the search allows the preferred Gaussian count to change after precision adaptation: at a given rate, a representation may favor more primitives with coarser attributes or fewer with finer precision. CP-GS therefore treats capacity and precision as exchangeable resources and jointly optimizes them rather than committing to either in advance.

\section{Experiments}

\subsection{Experimental Setup}
\label{sec:setup}

\paragraph{Datasets and metrics.}
We evaluate Gaussian representation fidelity and rate-distortion performance on DIV2K~\citet{timofte2017ntire} and Kodak using a unified protocol with fixed hyperparameters. Coding ablations are conducted on the DIV2K validation set and Kodak24. We report PSNR, SSIM~\citet{wang2004image}, MS-SSIM~\citet{wang2003multiscale}, and LPIPS~\citet{zhang2018unreasonable}, with the bitrate computed from complete serialized representations. The BD-rate\citet{HANHART2014555} is computed over valid overlapping quality intervals, and the number of valid images is reported for each comparison.

\paragraph{Implementation details.}
Experiments use 4 NVIDIA RTX 4090 GPUs. Unless otherwise stated, optimization time includes capacity reallocation, recovery optimization, precision calibration, serialization, decoding, and candidate evaluation. Content reallocation starts from a fixed Image-GS representation with 5,000 Gaussians after 50,000 optimization steps, followed by one reallocation and 10,000 recovery steps. Coding ablations use representations optimized for 120,000 iterations with at most 5,000 Gaussians. Precision is selected from $\{4,6,8,10,12,16\}$ under four payload budgets corresponding to uniform 6-, 8-, 10-, and 12-bit configurations, while count ablations retain 100\%, 80\%, 60\%, or 40\% of the fitted Gaussian set. 

\subsection{Comparison with Existing 2D Image Representations}

We compare CP-GS with representative INR- and GS-based image representation methods on Kodak and DIV2K. As shown in Table~\ref{tab:sota}, CP-GS achieves the best reconstruction quality among the compared methods at the evaluated Gaussian counts while maintaining compact parameterization and low optimization cost. Compared with its Image-GS backbone, CP-GS improves PSNR by 0.91 and 1.48\,dB on Kodak at 5K and 10K Gaussians, respectively, while reducing the corresponding single-image optimization time from 52.49 to 46.15\,s and from 71.31 to 54.90\,s. Similar gains are observed on DIV2K: CP-GS improves PSNR by 0.86\,dB at 10K Gaussians and 2.88\,dB at 30K, with higher MS-SSIM and lower optimization time than Image-GS. These results demonstrate that jointly optimizing Gaussian capacity and attribute precision can improve both representation fidelity and optimization efficiency.

\begin{table*}[t]
\centering
\vspace{-1em}
\caption{
Comparison with representative INR- and GS-based image representation methods on Kodak and DIV2K. GS-based methods are evaluated at 5K/10K Gaussians on Kodak and 10K/30K on DIV2K. We report PSNR, MS-SSIM, parameter count, and optimization time. Optimization time denotes the single-image wall-clock time required to obtain the final representation.
}
\label{tab:sota}
\setlength{\tabcolsep}{5pt}
\renewcommand{\arraystretch}{1.08}
\small

\resizebox{\textwidth}{!}{
\begin{tabular}{lcccc|cccc}
\toprule
\multirow{2}{*}{Methods}
&
\multicolumn{4}{c|}{Kodak}
&
\multicolumn{4}{c}{DIV2K}
\\
\cmidrule(lr){2-5}\cmidrule(lr){6-9}
&
PSNR
&
MS-SSIM
&
Params(M)
&
Optim. Time (s)
&
PSNR
&
MS-SSIM
&
Params(M)
&
Optim. Time (s)
\\
\midrule

\multicolumn{9}{l}{\textbf{INR-based}}\\

SIREN~\citet{sitzmann2020implicit}
& 26.50 & 0.8705 & 3.74 & 889
& 30.01 & 0.9630 & 0.42 & 2242
\\

SIREN+$T_{\mathrm{sym}}$~\citet{zhang2025enhancingimplicitneuralrepresentations}
& 32.80 & 0.9820 & 0.10 & 166
& 29.61 & 0.9712 & 0.10 & 261
\\

EVOS~\citet{Zhang_2025_CVPR}
& 33.64 & 0.9809 & 0.10 & 106
& 30.38 & 0.9687 & 0.10 & 189
\\
\midrule

\textbf{GS-based}
&
\multicolumn{4}{c|}{Number of Gaussians (5K)}
&
\multicolumn{4}{c}{Number of Gaussians (10K)}
\\
\cmidrule(lr){2-5}\cmidrule(lr){6-9}

GaussianImage~\citet{zhang2024gaussianimage}
& 29.94 & 0.9632 & 0.04 & 234.47
& 29.71 & 0.9727 & 0.08 & 260.12
\\

EA-GI~\citet{CHEN2026110521}
& 30.84 & 0.9614 & 0.04 & 247.80
& 30.23 & 0.9653 & 0.08 & 301.31
\\

GaussianImage++~\citet{li2026gaussianimagepp}
& 31.90 & 0.9614 & 0.04 & 247.75
& 32.05 & 0.9742 & 0.08 & 283.28
\\

Image-GS~\citet{zhang2025image}
& 36.50 & 0.9940 & 0.04 & 52.49
& 31.61 & 0.9684 & 0.08 & 74.37
\\

\textbf{CP-GS (Ours)}
& \textbf{37.41}
& \textbf{0.9950}
& 0.04
& \textbf{46.15}
& \textbf{32.47}
& \textbf{0.9771}
& 0.08
& \textbf{66.21}
\\
\midrule

& 
\multicolumn{4}{c|}{Number of Gaussians (10K)}
&
\multicolumn{4}{c}{Number of Gaussians (30K)}
\\
\cmidrule(lr){2-5}\cmidrule(lr){6-9}

GaussianImage~\citet{zhang2024gaussianimage}
& 32.58 & 0.9824 & 0.08 & 258.11
& 34.64 & 0.9926 & 0.24 & 287.15
\\

EA-GI~\citet{CHEN2026110521}
& 33.87 & 0.9806 & 0.08 & 267.21
& 35.59 & 0.9896 & 0.24 & 346.78
\\

GaussianImage++~\citet{li2026gaussianimagepp}
& 35.41 & 0.9832 & 0.08 & 283.20
& 38.30 & 0.9940 & 0.24 & 290.85
\\

Image-GS~\citet{zhang2025image}
& 39.31 & 0.9955 & 0.08 & 71.31
& 35.97 & 0.9913 & 0.24 & 112.43
\\

\textbf{CP-GS (Ours)}
& \textbf{40.79}
& \textbf{0.9958}
& 0.08
& \textbf{54.90}
& \textbf{38.85}
& \textbf{0.9943}
& 0.24
& \textbf{95.70}
\\
\bottomrule
\end{tabular}
}
\end{table*}

\begin{table*}[t]
\centering
\caption{Cumulative ablation of CP-GS on Kodak24.
Quality metrics are averaged over all 24 images, with changes relative to the Spatial baseline shown in parentheses. BD-rate is measured relative to 
Spatial.}
\label{tab:cumulative_ablation}
\setlength{\tabcolsep}{4.5pt}
\renewcommand{\arraystretch}{1.15}
\small
\resizebox{\textwidth}{!}{
\begin{tabular}{cccc|c|ccc}
\toprule
\multicolumn{4}{c|}{\textbf{CP-GS} Components} &
\multirow{2}{*}{\makecell{BD-Rate $\downarrow$\\
vs. \ Spatial (\%)}} &
\multicolumn{3}{c}{Quality Metrics}
\\
\cmidrule(lr){1-4}\cmidrule(lr){6-8}
\makecell{Spatial\\Realloc.} &
\makecell{Attr.\\RD} &
\makecell{Hier.\\Color} &
\makecell{Joint\\Capa.--Prec.} &
&
PSNR $\uparrow$ & SSIM $\uparrow$ & LPIPS $\downarrow$ \\
\midrule
\cmark & & & &
-- &
19.310 &
0.6457 &
0.3699
\\
\cmark & \cmark & & &
$-14.21$ &
22.740 ($+3.430$) &
0.6951 ($+0.0494$) &
0.3639 ($-0.0060$)
\\
\cmark & \cmark & \cmark & &
$-23.71$ &
23.225 ($+3.915$) &
0.7053 ($+0.0596$) &
0.3371 ($-0.0328$)
\\
\cmark & \cmark & \cmark & \cmark &
$-44.93$ &
\textbf{23.781} ($+4.471$) &
\textbf{0.7147} ($+0.0690$) &
\textbf{0.3308} ($-0.0392$)
\\
\bottomrule
\end{tabular}
}
\end{table*}

\subsection{Ablation Studies}
\label{sec:af-ablation}

We systematically analyze the three stages of CP-GS: content-adaptive 2D Gaussian representation, hierarchical distortion-aware representation quantization, and capacity-precision coupled optimization. Table~\ref{tab:cumulative_ablation} reports their cumulative contributions, while Table~\ref{tab:compact_ablation} provides targeted analyses of attribute precision allocation, hierarchical color refinement, and capacity-precision coupling. 

\subsubsection{Effect of Each Component}

Table~\ref{tab:cumulative_ablation} shows complementary gains from the three stages of CP-GS. Starting from the content-adaptive representation baseline, attribute-aware quantization reduces BD-rate by 14.21\% while improving PSNR by 3.43\,dB. Hierarchical color refinement further increases the BD-rate reduction to 23.71\%, while capacity-precision coupled optimization achieves the best overall quality and a 44.93\% BD-rate reduction. These progressive gains demonstrate that content-adaptive capacity utilization, hierarchical precision allocation, and their joint optimization provide complementary improvements in Gaussian representation efficiency.

\subsubsection{Effect of Content-Adaptive Gaussian Reallocation}

\begin{figure*}[t]
\vspace{-1em}
\centering
\includegraphics[width=\textwidth]{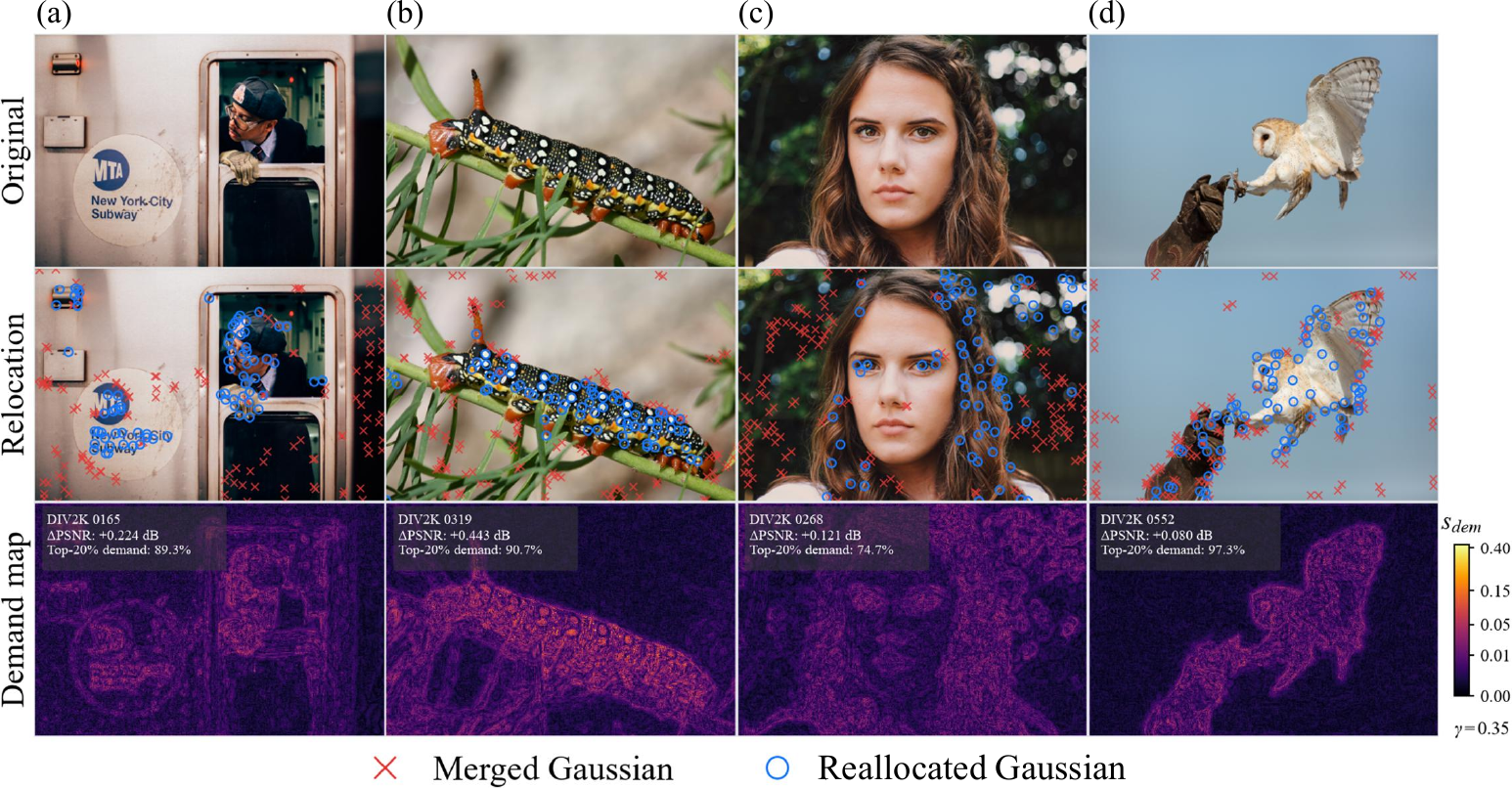}
\caption{Visualization of content-adaptive Gaussian capacity reallocation on four representative images (a)--(d). Rows show the original image, Gaussian reallocation, and the pre-reallocation demand map. Red crosses denote merged Gaussians, and blue circles denote newly reallocated Gaussians.}
\label{fig:spatialfig}
\end{figure*}

We evaluate whether demand-guided reallocation improves the utilization of a fixed Gaussian budget. All controls use the same merging and recovery settings and differ only in capacity placement. At $K=150$, demand-guided reallocation achieves 24.520\,dB PSNR, compared with 24.275\,dB for random placement and 24.211\,dB without reallocation, while placing 106 of the 150 newly allocated Gaussians in high-demand regions. Fig.~\ref{fig:spatialfig}(a)--(d) further shows that the reallocated capacity consistently follows high-demand structures. These results demonstrate that demand-guided reallocation improves capacity utilization without increasing the Gaussian budget.

\begin{table*}[t]
\centering
\vspace{-1em}
\caption{Ablation analysis of precision allocation and capacity-precision coupling. Physical BD-rate (\%, lower is better) is measured relative to Uniform in (a), Global in (b), and the fixed-count uniform baseline in (c). In (d), Structured-19 is compared with Oracle-216 under matched nominal payload targets; PSNR and BD-rate gaps are reported relative to Oracle-216.}
\label{tab:compact_ablation}

\setlength{\tabcolsep}{4.5pt}
\renewcommand{\arraystretch}{1.12}
\small

\resizebox{\linewidth}{!}{
\begin{tabular}{lcrr|lcrr}
\toprule

\multicolumn{4}{c|}{\textbf{(a) Attribute Precision Allocation}} &
\multicolumn{4}{c}{\textbf{(c) Capacity-Precision Coupling}} \\
\cmidrule(lr){1-4}\cmidrule(lr){5-8}

Method & Eval. & DIV2K (\%) & Kodak (\%) &
\multicolumn{2}{l}{Method} & DIV2K (\%) & Kodak (\%) \\
\midrule

Uniform (baseline)
& 0
& 0.00
& 0.00
&
\multicolumn{2}{l}{Baseline}
& 0.00
& 0.00
\\

Fixed mixed
& 0
& +21.35
& +20.65
&
\multicolumn{2}{l}{Hier. Precision}
& -7.18
& -7.59
\\

Greedy
& --
& -5.84
& -13.11
&
\multicolumn{2}{l}{Count-only}
& -32.91
& -48.12
\\

Random
& 16
& -8.47
& -3.19
&
\multicolumn{2}{l}{Precision-only}
& -17.00
& -14.30
\\

Oracle
& 216
& -17.94
& -14.33
&
\multicolumn{2}{l}{Joint RD}
& -45.01
& -50.14
\\

\textbf{Structured (Ours)}
& \textbf{19}
& \textbf{-17.00}
& \textbf{-14.30}
&
\multicolumn{2}{l}{\textbf{Joint Hier.\ RD (Ours)}}
& \textbf{-47.82}
& \textbf{-53.35}
\\

\midrule

\multicolumn{4}{c|}{\textbf{(b) Hierarchical Color Refinement}} &
\multicolumn{4}{c}{\textbf{(d) Structured-19 (Ours) vs.\ Oracle-216}} \\
\cmidrule(lr){1-4}\cmidrule(lr){5-8}

Method & Groups & DIV2K (\%) & Kodak (\%) &
\multicolumn{2}{l}{Metric} & DIV2K & Kodak \\
\midrule

Global
& 1
& 0.00
& 0.00
&
\multicolumn{2}{l}{Calibration evals.\ (S/O)}
& 19 / 216
& 19 / 216
\\

Shared
& 2
& -0.80
& -0.75
&
\multicolumn{2}{l}{Config.\ match (\%)}
& 87.50
& 93.75
\\

Random
& 2
& -1.31
& -1.62
&
\multicolumn{2}{l}{Mean PSNR gap (dB)}
& 0.038
& 0.032
\\

\textbf{Hierarchical (Ours)}
& 2
& \textbf{-7.09}
& \textbf{-10.68}
&
\multicolumn{2}{l}{BD-rate gap (\%)}
& 0.127
& 0.013
\\

\bottomrule
\end{tabular}
}
\end{table*}

\subsubsection{Ablation of Hierarchical Distortion-Aware Quantization}
\label{sec:af-attr}

We investigate attribute-level and intra-color precision allocation in Table~\ref{tab:compact_ablation}(a,b,d). Oracle-216 exhaustively evaluates the finite precision space and therefore provides an empirical upper bound for precision allocation. Structured-19 achieves physical BD-rate reductions of 17.00\% and 14.30\% on DIV2K and Kodak, respectively, closely approaching this upper bound with mean PSNR gaps of only 0.038 and 0.032\,dB. It requires approximately $11\times$ fewer logical calibration evaluations than Oracle-216 (19 vs.\ 216). Hierarchical K2 further reduces BD-rate by 7.09\% and 10.68\% over Global on DIV2K and Kodak, respectively, and consistently outperforms random grouping. These results show that distortion-aware precision allocation effectively captures quantization heterogeneity both across attributes and within color groups.

\vspace{-1em}
\subsubsection{Ablation of Capacity—Precision Coupling}

\begin{figure*}[t]
\centering
\includegraphics[width=0.75\textwidth]{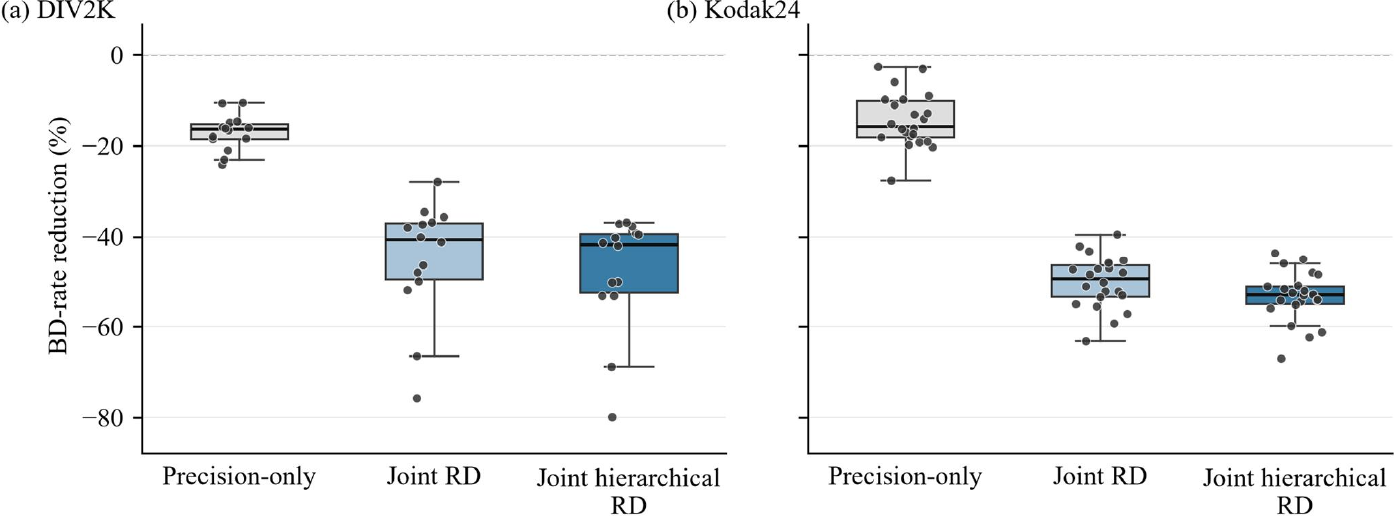}
\caption{Image-level BD-rate distributions of Precision-only, Joint RD, and Joint Hierarchical RD on (a) DIV2K and (b) Kodak24. Each point denotes one image, and the boxes summarize the corresponding distributions. A more negative BD-rate indicates a better rate-distortion efficiency.}
\label{fig:joint-box}
\end{figure*}

We further examine joint capacity—precision allocation under a shared budget. As shown in Table~\ref{tab:compact_ablation}(c), optimizing capacity or precision alone improves rate-distortion efficiency, while Joint RD achieves larger BD-rate reductions of 45.01\% and 50.14\% on DIV2K and Kodak, respectively. Adding hierarchical precision further improves these reductions to 47.82\% and 53.35\%, respectively. Fig.~\ref{fig:joint-box} shows the same overall trend across individual images, demonstrating that capacity and precision are complementary resources that benefit from joint optimization.

\vspace{-1em}

\section{Conclusion}

We presented CP-GS, a capacity-precision coupled framework for budget-aware 2D Gaussian image representation. CP-GS improves Gaussian capacity utilization through content-adaptive reallocation, assigns precision according to hierarchical distortion sensitivity, and jointly optimizes capacity and precision under a shared budget. Experiments on DIV2K and Kodak demonstrate strong rate-distortion performance, validating the benefit of jointly considering where Gaussian capacity is allocated and how precisely it is encoded.

\noindent \textbf{Broader Implications and Limitations.}
The capacity-precision coupling explored in CP-GS may generalize to other resource-constrained explicit representations. However, CP-GS currently relies on image-specific calibration and a predefined candidate space, while its evaluation is limited to static 2D images. Future work will explore learned allocation strategies, more efficient rate-distortion search, and broader visual representation tasks.





\bibliography{XX2027_conference}

@inproceedings{zhang2024gaussianimage,
  title={GaussianImage: 1000 FPS Image Representation and Compression by 2D Gaussian Splatting},
  author={Zhang, Xinjie and Ge, Xingtong and Xu, Tongda and He, Dailan and Wang, Yan and Qin, Hongwei and Lu, Guo and Geng, Jing and Zhang, Jun},
  booktitle={European Conference on Computer Vision (ECCV)},
  pages={327--345},
  year={2024},
  publisher={Springer Nature Switzerland}
}

@misc{chen2026locoadc,
      title={Locality-Aware Density Control for Efficient Gaussian-based Image Representation}, 
      author={Jiacong Chen and Qingyu Mao and Xiandong Meng and     Shuai Liu and Chao Li and Fanyang Meng and Youneng Bao and Yongsheng Liang},
      year={2026},
      eprint={2607.17896},
      journal={arXiv preprint arXiv:2607.17896},
      primaryClass={cs.CV},
}

@inproceedings{pan2026sgi,
    author    = {Pan, Zixuan and Tang, Kaiyuan and Xia, Jun and Qin, Yifan and Gu, Lin and Wang, Chaoli and Chen, Jianxu and Shi, Yiyu},
    title     = {SGI: Structured 2D Gaussians for Efficient and Compact Large Image Representation},
    booktitle = {Proceedings of the IEEE/CVF Conference on Computer Vision and Pattern Recognition (CVPR)},
    month     = {June},
    year      = {2026},
    pages     = {12162-12172}
}

@inproceedings{peng2025continuoussr,
 author = {Long Peng and Anran Wu and Wenbo Li and Peizhe Xia and Xueyuan Dai and Xinjie Zhang and Xin Di and Haoze Sun and Renjing Pei and Yang Wang and Yang Cao and Zheng-Jun Zha},
 booktitle = {International Conference on Learning Representations (ICLR)}, 
 pages = {92172--92199},
 title = {Pixel to Gaussian: Ultra-Fast Continuous Super-Resolution with 2D Gaussian Modeling},
 volume = {2026},
 year = {2026}
}

@inproceedings{li2026gaussianimagepp,
  title     = {{GaussianImage++}: Boosted Image Representation and Compression with {2D} {Gaussian} Splatting},
  author    = {Li, Tiantian and Zhang, Xinjie and Ge, Xingtong and Xu, Tongda and He, Dailan and Zhang, Jun and Wang, Yan},
  booktitle = {Proceedings of the AAAI Conference on Artificial Intelligence (AAAI)},
  volume    = {40},
  pages     = {6442--6449},
  year      = {2026}
}

@inproceedings{zhang2025image,
  title={Image-gs: Content-adaptive image representation via 2d gaussians},
  author={Zhang, Yunxiang and Li, Bingxuan and Kuznetsov, Alexandr and Jindal, Akshay and Diolatzis, Stavros and Chen, Kenneth and Sochenov, Anton and Kaplanyan, Anton and Sun, Qi},
  booktitle={Proceedings of the Special Interest Group on Computer Graphics and Interactive Techniques Conference Papers},
  pages = {102:1--102:11},
  publisher = {Association for Computing Machinery},
  year={2025}
}

@article{zhang2025enhancingimplicitneuralrepresentations,
  author = {Weixiang Zhang and Shuzhao Xie and Chengwei Ren and Shijia Ge and Mingzi Wang and Zhi Wang},
  title = {Enhancing Implicit Neural Representations via Symmetric Power Transformation},
  journal = {Proceedings of the AAAI Conference on Artificial Intelligence (AAAI)},
  volume = {39},
  number = {10},
  pages = {10157--10165},
  year = {2025}
}

@inproceedings{zhang20252dmamba,
  title={2Dmamba: Efficient state space model for image representation with applications on giga-pixel whole slide image classification},
  author={Zhang, Jingwei and Nguyen, Anh Tien and Han, Xi and Trinh, Vincent Quoc-Huy and Qin, Hong and Samaras, Dimitris and Hosseini, Mahdi S},
  booktitle={Proceedings of the IEEE/CVF Conference on Computer Vision and Pattern Recognition (CVPR)},
  pages={3583--3592},
  year={2025},
  organization={IEEE}
}

@inproceedings{chen2021learning,
  title={Learning continuous image representation with local implicit image function},
  author={Chen, Yinbo and Liu, Sifei and Wang, Xiaolong},
  booktitle={Proceedings of the IEEE/CVF Conference on Computer Vision and Pattern Recognition (CVPR)},
  pages={8628--8638},
  year={2021}
}

@inproceedings{kao2025bridging,
  title={Bridging compressed image latents and multimodal large language models},
  author={Kao, Chia-Hao and Chien, Cheng and Tseng, Yu-Jen and Chen, Yi-Hsin and Gnutti, Alessandro and Lo, Shao-Yuan and Peng, Wen-Hsiao and Leonardi, Riccardo},
  booktitle={Proceedings of the International Conference on Learning Representations (ICLR)},
  volume={2025},
  pages={7966--7986},
  year={2025}
}

@inproceedings{xiao2025omnigen,
  title={Omnigen: Unified image generation},
  author={Xiao, Shitao and Wang, Yueze and Zhou, Junjie and Yuan, Huaying and Xing, Xingrun and Yan, Ruiran and Li, Chaofan and Wang, Shuting and Huang, Tiejun and Liu, Zheng},
  booktitle={Proceedings of the IEEE/CVF Conference on Computer Vision and Pattern Recognition (CVPR)},
  pages={13294--13304},
  year={2025},
  organization={IEEE}
}

@inproceedings{tian2025degradation,
  title={Degradation-aware feature perturbation for all-in-one image restoration},
  author={Tian, Xiangpeng and Liao, Xiangyu and Liu, Xiao and Li, Meng and Ren, Chao},
  booktitle={Proceedings of the IEEE/CVF Conference on Computer Vision and Pattern Recognition (CVPR)},
  pages={28165--28175},
  year={2025},
  organization={IEEE}
}

@article{jiang2025survey,
  title={A survey on all-in-one image restoration: Taxonomy, evaluation and future trends},
  author={Jiang, Junjun and Zuo, Zengyuan and Wu, Gang and Jiang, Kui and Liu, Xianming},
  journal={IEEE Transactions on Pattern Analysis and Machine Intelligence},
  volume = {47},
  number = {12},
  pages = {11892--11911},
  year={2025},
  publisher={IEEE}
}

@inproceedings{ye2026imgedit,
  title     = {{ImgEdit}: A Unified Image Editing Dataset and Benchmark},
  author    = {Ye, Yang and He, Xianyi and Li, Zongjian and Yuan, Shenghai and Yan, Zhiyuan and Hou, Bohan and Yuan, Li and others},
  booktitle = {Proceedings of the Conference on Neural Information Processing Systems (NeurIPS)},
  volume    = {38},
  year      = {2026}
}

@inproceedings{sitzmann2020implicit,
  title     = {Implicit Neural Representations with Periodic Activation Functions},
  author    = {Sitzmann, Vincent and Martel, Julien N. P. and Bergman, Alexander W. and Lindell, David B. and Wetzstein, Gordon},
  booktitle = {Proceedings of the Conference on Neural Information Processing Systems (NeurIPS)},
  volume    = {33},
  pages     = {7462--7473},
  year      = {2020}
}

@article{dupont2021coin,
  title={Coin: Compression with implicit neural representations},
  author={Dupont, Emilien and Goli{\'n}ski, Adam and Alizadeh, Milad and Teh, Yee Whye and Doucet, Arnaud},
  journal={arXiv preprint arXiv:2103.03123},
  year={2021},
  eprint = {2103.03123}
}

@inproceedings{hu2025gaussiansr,
  title     = {{GaussianSR}: High Fidelity {2D} {Gaussian} Splatting for Arbitrary-Scale Image Super-Resolution},
  author    = {Hu, Jintong and Xia, Bin and Chen, Bin and Yang, Wenming and Zhang, Lei},
  booktitle = {Proceedings of the AAAI Conference on Artificial Intelligence (AAAI)},
  volume    = {39},
  pages     = {3554--3562},
  year      = {2025}
}

@inproceedings{zeng2025instant,
  author = {Zeng, Zhaojie and Wang, Yuesong and Guan, Tao and Yang, Chao and Ju, Lili},
  title = {Instant {GaussianImage}: A Generalizable and Self-Adaptive Image Representation via {2D} {Gaussian} Splatting},
  booktitle = {Proceedings of the IEEE/CVF International Conference on Computer Vision (ICCV)},
  pages = {27896--27905},
  year = {2025},
}

@inproceedings{wan2025spectrum,
  title     = {Spectrum-Adaptive Distribution of {2D} {Gaussians} for Image Representation and Compression},
  author    = {Wan, Zunian and Zhao, Jiancheng and Ding, Yepeng and Zhang, Lingfeng and Sato, Hiroyuki and Ogawa, Takefumi},
  booktitle = {Proceedings of the IEEE International Conference on Multimedia and Expo (ICME)},
  pages     = {1--6},
  year      = {2025}
}

@inproceedings{zhang2026gpstoken,
  title     = {{GPSToken}: {Gaussian} Parameterized Spatially-Adaptive Tokenization for Image Representation and Generation},
  author    = {Zhang, Zhengqiang and Wu, Rongyuan and Sun, Lingchen and Zhang, Lei},
  booktitle = {Proceedings of the Conference on Neural Information Processing Systems (NeurIPS)},
  volume    = {38},
  pages     = {11218--11249},
  year      = {2026}
}

@article{chen2026sparegs,
  title={SPARE-GS: Structural Parsimony and Resource Efficiency for 3D Gaussian Splatting},
  author={Chen, Zhang and Wan, Shuai and Yang, Fuzheng and Xia, Jiazhi and Lin, Weiyao and Hou, Junhui},
  journal={arXiv preprint arXiv:2607.16624},
  eprint = {2607.16624},
  year={2026}
}

@inproceedings{zhou2026gdags,
  title     = {Gradient-Direction-Aware Density Control for {3D} {Gaussian} Splatting},
  author    = {Zhou, Zheng and Xiong, Yu-Jie and Zhang, Jia-Chen and Xia, Chun-Ming and Qiu, Xihe and Zhan, Hongjian},
  booktitle = {Proceedings of the International Conference on Learning Representations (ICLR)},
  volume    = {2026},
  pages     = {21415--21437},
  year      = {2026}
}

@article{liang2025structure,
  title={Structure-Guided Allocation of {2D} {Gaussians} for Image Representation and Compression},
  author={Liang, Huanxiong and Chen, Yunuo and Pan, Yicheng and Wang, Sixian and Dai, Jincheng and Lu, Guo and Zhang, Wenjun},
  journal={arXiv preprint arXiv:2512.24018},
    eprint = {2512.24018},
  year={2025}
}

@article{cheng2026cgvq,
  title={Clustered Codebook Quantization for 2D Gaussian-based Image Compression},
  author={Cheng, Runze and Zhan, Yicheng and Spjut, Josef and Ak{\c{s}}it, Kaan},
  journal={arXiv preprint arXiv:2607.05667},
  year={2026}
}

@inproceedings{wang2024rdogaussian,
  title     = {End-to-End Rate-Distortion Optimized {3D} {Gaussian} Representation},
  author    = {Wang, Henan and Zhu, Hanxin and He, Tianyu and Feng, Runsen and Deng, Jiajun and Bian, Jiang and Chen, Zhibo},
  booktitle = {Proceedings of the European Conference on Computer Vision (ECCV)},
  pages     = {76--92},
  year      = {2024}
}

@inproceedings{zhan2025cat3dgs,
  title     = {{CAT-3DGS}: A Context-Adaptive Triplane Approach to Rate-Distortion-Optimized {3DGS} Compression},
  author    = {Zhan, Yu-Ting and Ho, Cheng-Yuan and Yang, He-Bi and Chen, Yi-Hsin and Chiang, Jui-Chiu and Liu, Yu-Lun and Peng, Wen-Hsiao},
  booktitle = {Proceedings of the International Conference on Learning Representations (ICLR)},
  volume    = {2025},
  pages     = {26312--26334},
  year      = {2025}
}

@inproceedings{tseng2026csgaussian,
  title     = {{CSGaussian}: Progressive Rate-Distortion Compression and Segmentation for {3D} {Gaussian} Splatting},
  author    = {Tseng, Yu-Jen and Kao, Chia-Hao and Chen, Jing-Zhong and Gnutti, Alessandro and Lo, Shao-Yuan and Lin, Yen-Yu and Peng, Wen-Hsiao},
  booktitle = {Proceedings of the IEEE/CVF Winter Conference on Applications of Computer Vision (WACV)},
  pages     = {6883--6892},
  year      = {2026}
}

@inproceedings{chen2024hac,
  title     = {{HAC}: Hash-Grid Assisted Context for {3D} {Gaussian} Splatting Compression},
  author    = {Chen, Yihang and Wu, Qianyi and Lin, Weiyao and Harandi, Mehrtash and Cai, Jianfei},
  booktitle = {Proceedings of the European Conference on Computer Vision (ECCV)},
  pages     = {422--438},
  year      = {2024}
}

@inproceedings{chen2026pcgs,
  title     = {{PCGS}: Progressive Compression of {3D} {Gaussian} Splatting},
  author    = {Chen, Yihang and Li, Mengyao and Wu, Qianyi and Lin, Weiyao and Harandi, Mehrtash and Cai, Jianfei},
  booktitle = {Proceedings of the AAAI Conference on Artificial Intelligence (AAAI)},
  volume    = {40},
  pages     = {3111--3119},
  year      = {2026}
}

@inproceedings{huang2025hilcomp,
  title     = {{HilComp}: Hilbert Curve-Based Balanced Clustering for {3D} {Gaussian} Splatting Compression},
  author    = {Huang, Yu and Fu, Xiaowen and Ma, Ling and He, Zhengqi and Wang, Jinbao and Li, Xueliang},
  booktitle = {Proceedings of the International Conference on Intelligent Computing (ICIC)},
  pages     = {421--432},
  year      = {2025}
}

@InProceedings{timofte2017ntire,
author = {Agustsson, Eirikur and Timofte, Radu},
title = {NTIRE 2017 Challenge on Single Image Super-Resolution: Dataset and Study},
booktitle = {Proceedings of the IEEE Conference on Computer Vision and Pattern Recognition (CVPR) Workshops},
month = {July},
year = {2017}
}

@misc{kodak,
  title={Kodak Lossless True Color Image Suite},
  author={Kodak.},
  year={1999},
}

@ARTICLE{wang2004image,
  author={Zhou Wang and Bovik, A.C. and Sheikh, H.R. and Simoncelli, E.P.},
  journal={IEEE Transactions on Image Processing}, 
  title={Image quality assessment: from error visibility to structural similarity}, 
  year={2004},
  volume={13},
  number={4},
  pages={600-612},
}

@INPROCEEDINGS{wang2003multiscale,
  author={Wang, Z. and Simoncelli, E.P. and Bovik, A.C.},
  booktitle={The Thrity-Seventh Asilomar Conference on Signals, Systems \& Computers, 2003}, 
  title={Multiscale structural similarity for image quality assessment}, 
  year={2003},
  volume={2},
  number={},
  pages={1398-1402 Vol.2},
}

@InProceedings{zhang2018unreasonable,
author = {Zhang, Richard and Isola, Phillip and Efros, Alexei A. and Shechtman, Eli and Wang, Oliver},
title = {The Unreasonable Effectiveness of Deep Features as a Perceptual Metric},
booktitle = {Proceedings of the IEEE Conference on Computer Vision and Pattern Recognition (CVPR)},
month = {June},
year = {2018}
}

@article{HANHART2014555,
title = {Calculation of average coding efficiency based on subjective quality scores},
journal = {Journal of Visual Communication and Image Representation},
volume = {25},
number = {3},
pages = {555-564},
year = {2014},
note = {QoE in 2D/3D Video Systems},
issn = {1047-3203},
author = {Philippe Hanhart and Touradj Ebrahimi},
}

@article{CHEN2026110521,
title = {Entropy-aware image representation via 2D Gaussian splatting},
journal = {Signal Processing},
volume = {244},
pages = {110521},
year = {2026},
issn = {0165-1684},
author = {Jiacong Chen and Qingyu Mao and Shuai Liu and Chao Li and Jierun Lin and Xiandong Meng and Fanyang Meng and Yongsheng Liang},
}

@InProceedings{Zhang_2025_CVPR,
    author    = {Zhang, Weixiang and Xie, Shuzhao and Ren, Chengwei and Xie, Siyi and Tang, Chen and Ge, Shijia and Wang, Mingzi and Wang, Zhi},
    title     = {EVOS: Efficient Implicit Neural Training via EVOlutionary Selector},
    booktitle = {Proceedings of the IEEE/CVF Conference on Computer Vision and Pattern Recognition (CVPR)},
    month     = {June},
    year      = {2025},
    pages     = {30472-30482}
}
\bibliographystyle{XX2027_conference}

\end{document}